\documentclass[10pt,twocolumn,letterpaper]{article}

\usepackage{iccv}
\usepackage{times}
\usepackage{epsfig}
\usepackage{graphicx}
\usepackage{amsmath}
\usepackage{amssymb}
\usepackage{xcolor}
\usepackage{multirow}
\usepackage{cite}
\usepackage{bbold}

\DeclareMathOperator*{\argmin}{argmin}

\usepackage[pagebackref=true,breaklinks=true,letterpaper=true,colorlinks,bookmarks=false]{hyperref}

\iccvfinalcopy 

\def\iccvPaperID{1092} 

\ificcvfinal\fi
\begin{document}

\title{Zero-Shot Grounding of Objects from Natural Language Queries}

\author{Arka Sadhu$^1$ \quad \quad Kan Chen$^{2\dagger}$ \quad \quad Ram Nevatia$^1$\\
$^1$University of Southern California \quad \quad $^2$Facebook Inc.\\
{\tt\small {\{asadhu|nevatia\}@usc.edu} \quad kanchen18@fb.com}
}

\maketitle
\ificcvfinal\thispagestyle{empty}\fi

\ificcvfinal
\renewcommand*{\thefootnote}{$\dagger$}
\footnotetext{This work was done while the author was at USC.}
\renewcommand*{\thefootnote}{\arabic{footnote}}
\fi

\begin{abstract}
    A phrase grounding system localizes a particular object in an image referred to by a natural language query. In previous work, the phrases were restricted to have nouns that were encountered in training, we extend the task to \textbf{Zero-Shot Grounding}(\textbf{ZSG}) which can include novel, ``unseen" nouns. Current phrase grounding systems use an explicit object detection network in a 2-stage framework where one stage generates sparse proposals and the other stage evaluates them. In the ZSG setting, generating appropriate proposals itself becomes an obstacle as the proposal generator is trained on the entities common in the detection and grounding datasets. We propose a new single-stage model called \textbf{ZSGNet} which combines the detector network and the grounding system and predicts classification scores and regression parameters. Evaluation of ZSG system brings additional subtleties due to the influence of the relationship between the query and learned categories; we define four distinct conditions that incorporate different levels of difficulty. We also introduce new datasets, sub-sampled from Flickr30k Entities and Visual Genome, that enable evaluations for the four conditions. Our experiments show that  ZSGNet achieves state-of-the-art performance on Flickr30k and ReferIt under the usual ``seen" settings and performs significantly better than baseline in the zero-shot setting.
\end{abstract}

\section{Introduction}

Detecting objects in an image is a fundamental objective of computer vision. A variation of this task is \textit{phrase grounding} (also called \textit{visual grounding} and \textit{referring expressions}) where the objective is to detect objects referenced by noun phrases in a text query \cite{hu2016natural,rohrbach2016grounding,chen2017query, yu2018mattnet}.
It can be directly applied to other tasks such as visual question answering \cite{antol2015vqa, zhang2018interpretable} and image retrieval \cite{chen2017amc} and has thus garnered wide interest.

\begin{figure}[t]
    \centering
    \begin{tabular}{ccc}
        \includegraphics[width=0.27\linewidth, height=0.27\linewidth]{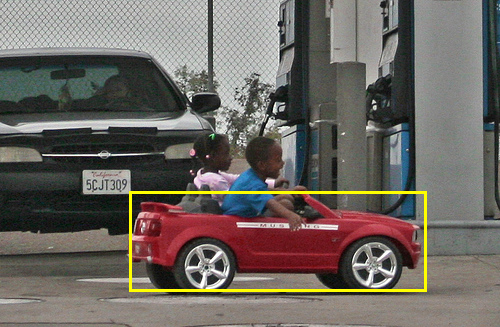} &  \includegraphics[width=0.27\linewidth, height=0.27\linewidth]{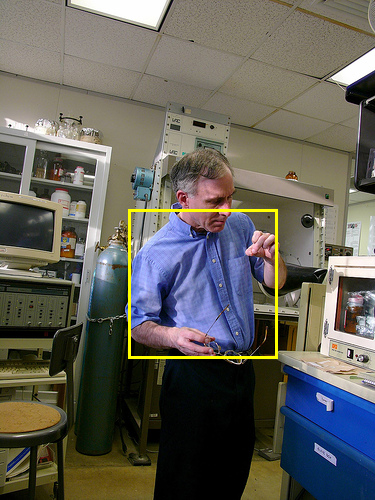} &
        \includegraphics[width=0.27\linewidth, height=0.27\linewidth]{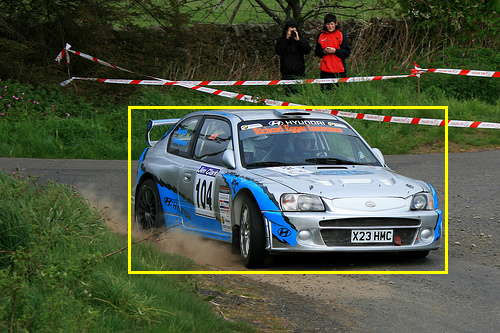}\\
        (a) red \textit{car} & (b) blue \textit{shirt} & (c) blue \textit{car}\\
         \includegraphics[width=0.27\linewidth, height=0.27\linewidth]{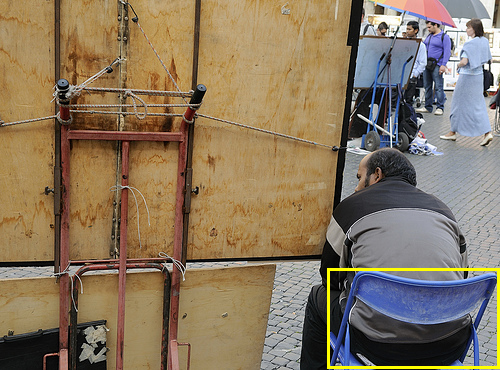} &
        \includegraphics[width=0.27\linewidth, height=0.27\linewidth]{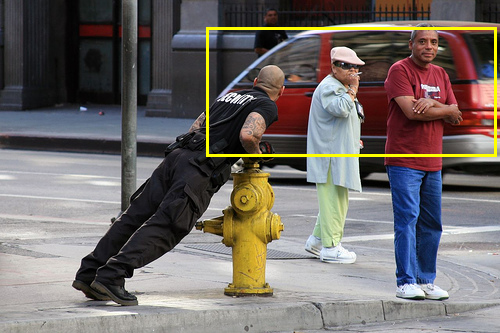} &  \includegraphics[width=0.27\linewidth, height=0.27\linewidth]{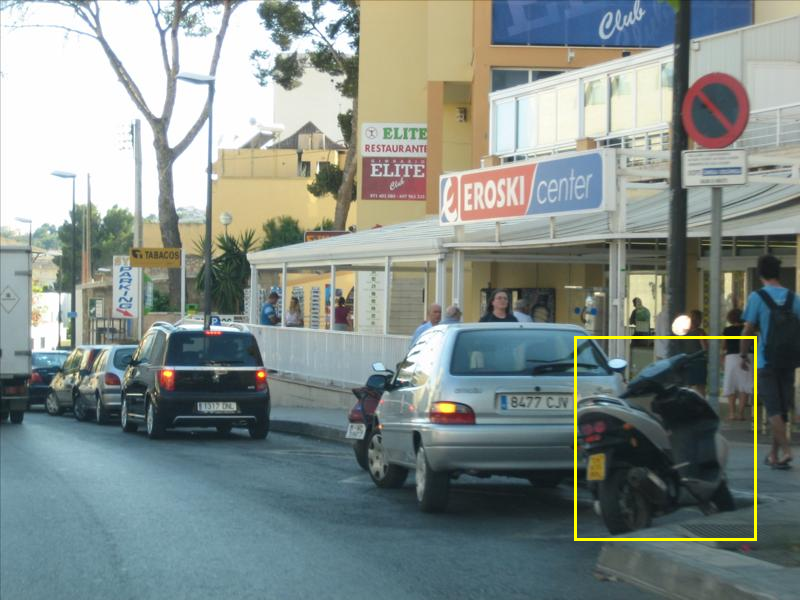}\\
         (d) blue \textit{chair} & (e) red \textit{minivan} & (f) silver \textit{moped} \\
    \end{tabular}
    \caption{Illustration of the key difference between current scope of phrase grounding and the proposed zero-shot grounding. The query word is italicized in all cases. (a)-(f) denote the image-query pairs input to the system. (a) and (b) are  examples of training images. A  test image query pair for phrase grounding could be (c). Zero-shot grounding additionally can be tested on (d), (e) and (f) in which ``chair'', ``minivan'' and ``moped'' are object categories not annotated in the training data. (f) additionally contains a  ``car'' object which is a trained category, indicating that both novel and related trained category objects may be present in a test image.}
    \label{fig:example_imgs}
\end{figure}

While existing phrase grounding systems accept novel query phrases as inputs, they are limited to the nouns encountered in the training data (\ie the referred object types need to have been ``seen" in training images before).
As an important extension, we define zero-shot grounding (\textbf{ZSG}) to allow the use of phrases with  nouns that the grounding system has not encountered in training set before.
Fig \ref{fig:example_imgs} illustrates this concept with examples.

To enable grounding of novel object categories, we need to relate the appearance of referred objects to their linguistic descriptions.
Current, state-of-art phrase grounding systems \cite{yu2018mattnet, deng2018visual, chen2017query, yu2018rethinking, plummer2018conditional} rely on an explicit object detector to obtain proposed object bounding boxes and their ROI-pooled features as a pre-processing step.
This essentially limits these systems to a fixed set of object categories that the detector was trained on. In ZSG,  we need to have a reasonable proposal box for the novel object, classify it as a foreground and regress the box to be a more accurate spatial fit.
In traditional phrase grounding, a key challenge is to disambiguate between similar objects using the query phrase, but ZSG  requires us to also first find likely image regions that may contain the referenced objects.

To address the above issues, we replace the traditional two-stage approach, where the first stage generates proposal bounding boxes and the second stage does the classification, by a single-stage network with dense proposals; we call this network \textbf{ZSGNet}.
It takes combined language query features and visual features from the image proposals and predicts classification scores and regression parameters.
The system is trained directly on the grounding training data, in an end-to-end manner, and does not utilize any externally trained object detector.
We show that, besides enabling grounding of novel categories, it does not degrade performance on learned categories even though our method does not utilize external training data. Moreover, our design is computationally efficient especially during inference owing to its single-stage architecture akin to SSD \cite{liu2016ssd}.

Evaluating the performance of a ZSG method is complex due to the influence of the relationship of the new query category to the learned categories.
To make the evaluations and distinctions clearer, we define four specific cases for different conditions:
(i) when the query word is novel (Fig \ref{fig:example_imgs} d-f)
(ii) when the referred object
belongs to a novel category  (Fig \ref{fig:example_imgs}-d)
(iii)
when the referred object is ``similar'' to
objects seen during training
but none of the latter are present (Fig \ref{fig:example_imgs}-e)
(iv) when at least one similar object also exists in the test image (Fig \ref{fig:example_imgs}-f)(more details in Section \ref{s:zsg_exp}).

To support evaluation of zero-shot grounding for the four cases, we introduce new datasets which are sub-sampled from the existing Visual Genome \cite{krishna2017visual} and Flickr30k Entities \cite{plummer2015flickr30k}.
We create examples of the four cases outlined above (dataset creation details are in Section \ref{s:dc}, experiments on these datasets are in Section \ref{r:unseen}).

Our contributions can be summarized as follows:
(i) we introduce the problem of Zero-shot grounding,
(ii) we propose a simple yet effective architecture ZSGNet to address limitations of current phrase grounding systems for this task,
(iii) we create new datasets suitable for evaluating zero-shot grounding and
(iv) we evaluate  performance on these datasets and show the effectiveness of our approach.
Our code and datasets are publicly released\footnote{\url{https://github.com/TheShadow29/zsgnet-pytorch}}.

\section{Related Work}
\label{s:related}

\textbf{Phrase grounding:} Extensive work in creating grounding datasets like Flickr30k, ReferIt, RefCoCo, RefCoCo+, RefCoCog, Visual Genome, GuessWhat \cite{plummer2015flickr30k, KazemzadehOrdonezMattenBergEMNLP14, yu2016modeling, mao2016generation, krishna2017visual, de2017guesswhat, flickr30k} have been crucial to the success of phrase grounding. Early works use reconstruction based approach \cite{rohrbach2016grounding} or integrate global context with the spatial configurations \cite{hu2016natural}.
Recent approaches \cite{chen2017query, yu2018rethinking,plummer2018conditional} learn directly in the  multi-modal feature space and use attention mechanisms \cite{yu2018mattnet, deng2018visual} which have also been extended to phrase grounding in dialogue systems \cite{de2017guesswhat, zhuang2018parallel}.
Few approaches also look at unsupervised learning
using variational context \cite{zhang2018grounding} and semi-supervised learning via gating mechanisms \cite{chen2018knowledge}.

Above techniques use an object detector like FasterR-CNN \cite{ren2015faster} or MaskR-CNN \cite{he2017mask} as a pre-processing step to get the bounding boxes and ROI-pooled features which effectively limits them to the object categories of the detector.
We combine the detection and the grounding networks and learn directly from the grounding dataset and thus no pre-processing step is involved.

\textbf{Multi-modal feature} representation has many flavors like linear transformation, concatenation, hadamard product \cite{kim2016hadamard}, bilinear pooling \cite{lin2015bilinear} and have shown success in vision-language tasks like VQA \cite{fukui2016multimodal, yu2018beyond, yu2017multi, ben2017mutan, ben2019block}, Scene Graph Generations\cite{li2017scene, yang2018graph} and Image Captioning \cite{lu2018neural}. We stick to feature concatenation for simplicity and a fair comparison with previous works in phrase grounding.

\textbf{Single stage networks} used in object detection are popular for their real-time inference speed. Prominent works include SSD \cite{liu2016ssd}, YOLO \cite{redmon2016you, redmon2017yolo9000, redmon2018yolov3} and more recently FPN \cite{lin2017feature} and RetinaNet \cite{lin2018focal}. In this work, we combine a single-stage detection network directly into the grounding framework; besides enabling zero-shot grounding, it also results in highly efficient inference.

\textbf{Zero-shot grounding} is unexplored but there are a few similar works. \cite{guadarrama2014open} aims at open-vocabulary object retrieval though it still assumes the entity of the referred object is seen at train time. Recently \cite{bansal2018zero} proposed zero-shot detection (ZSD) where they consider a set of unseen classes for which no bounding box information at train time. At test time, all the objects including the unseen classes must be detected.
However, the set of background classes is needed prior to training, but this is not needed in ZSG.

\section{Design Considerations for ZSG}
\label{s:zsg}
We fist discuss the zero-shot grounding cases and then describe the limitations in extending current phrase grounding systems to ZSG. Finally, we present a new architecture to address the limitations.

\subsection{ZSG Cases}
\label{s:zsg_exp}

\begin{table}[t]
\centering
\begin{tabular}{|c|c|c|}
\hline
Notation & Meaning & Example \\ \hline
$T$ & Test Image & Fig \ref{fig:example_imgs}-(f) \\ \hline
$P$ & Test query phrase & silver moped \\ \hline
$A$ & Referred object at test time & Moped \\ \hline
$Q$ & Word in $P$ referring to $A$ & moped \\ \hline
$B$ & \begin{tabular}[c]{@{}c@{}}Set of objects close to $A$ \\and seen during training\end{tabular}& \{Car\} \\ \hline
$C$ & \begin{tabular}[c]{@{}c@{}}Set of categories  \\ seen during training\end{tabular} & \begin{tabular}[c]{@{}c@{}}\{Vehicles, \\ Clothing\}\end{tabular}\\ \hline
$W$ & \begin{tabular}[c]{@{}c@{}}Set of words seen \\ during training\end{tabular} & \begin{tabular}[c]{@{}c@{}}\{red, blue, \\ car, shirt\}\end{tabular} \\ \hline
\end{tabular}
\vspace{2mm}
\caption{Notations used to describe ZSG with examples (Fig \ref{fig:example_imgs}). By close objects we mean their word embeddings are similar.}
\label{t:notation}
\end{table}

We now describe the four cases for zero-shot grounding in detail.
For brevity, we use the notations in Table \ref{t:notation}.
Each case defines the scope for what is classified as a zero-shot grounding example.
Further, we assume that $Q$ (the word which refers to the object in the image) is not an OOV (out of vocabulary word) which is reasonable if we use word embeddings which are trained on a large language corpus.

\noindent \textbf{Case 0:} $Q \notin W$. The query noun, $Q$, is not included in any training example before. We only look at the lemmatized word so synonyms are considered to be different (novel) words. Fig \ref{fig:example_imgs}(d)-(f) are  examples of this case.  Fig \ref{fig:example_imgs}-c with the phrase ``blue automobile'' would also be considered zero-shot since we haven't seen the word automobile before even though it is a synonym of  ``car''.\\
\noindent \textbf{Case 1:}
$A \notin C$.
Here, we assume that objects seen at train time belong to a set of pre-defined categories
and the referred object $A$ doesn't belong to these categories.
In  Fig \ref{fig:example_imgs}-d,   ``chair''  is considered zero-shot as this category was not seen at train time. \\
\noindent \textbf{Case 2:}
$\exists B$ but $\forall b \in B$ we have $b \notin T$.
Here, objects that are semantically close (similar) to the referred object $A$ are present in the training set but not in the test image.
Fig \ref{fig:example_imgs}-e is an example as ``minivan'' (novel object) is semantically close to ``car'' (seen in train set) but there is no other similar object like ``car'' in the test image.\\
\noindent \textbf{Case 3:}
$\exists B$ and $\exists b \in B$ such that $b \in T$.
Same as Case 2 but at least one of the objects semantically close (similar) to $A$ is also present in the test image.
For example, Fig \ref{fig:example_imgs}-f containing ``moped'' (a novel object) and ``car'' (seen in the training set) which are semantically close.

For Case 2 and Case 3, there can be multiple interpretations for being ``semantically close''. Here, we assume two objects are ``close'' if their word embeddings are similar. In our implementation, we cluster the word embeddings of the objects and objects belonging to the same cluster are considered semantically close (more details in Section \ref{s:dc}).

\subsection{Limitations in Phrase Grounding Systems}
Prior works view phrase grounding either as an entity selection problem \cite{yu2018mattnet, deng2018visual} or that of regressing sparse proposals to a tighter bounding box \cite{chen2017query, yu2018rethinking, plummer2018conditional}. In either case, given an image $I$, we  have $N$ candidate boxes and their ROI-pooled features $\{o_i\}_{i=1}^N$.
 Given a query phrase $P$, the problem reduces to finding a good candidate box with a possible additional regression step.

Grounding systems using this framework don't have a mechanism to generalize to object categories not in the detection dataset.
Consider a novel category $X$ whose instances may be present in the training images but not annotated. The  object detector learns to classify $X$ as background and this error is propagated to the grounding system.
\cite{chen2017query, plummer2018conditional} suggest fine-tuning the detector on the grounding categories (entities) but the grounding datasets are not densely annotated, \ie not all object instances of $X$ in every image are annotated. Additionally, some
  grounding datasets like ReferIt \cite{KazemzadehOrdonezMattenBergEMNLP14} don't have entity information so fine-tuning is not feasible.

  Object detectors also favor features invariant to intra-class changes but grounding systems need to capture intra-class variations as well.

\subsection{Model Design}
\begin{figure}[t]
    \centering
    \includegraphics[width=\linewidth]{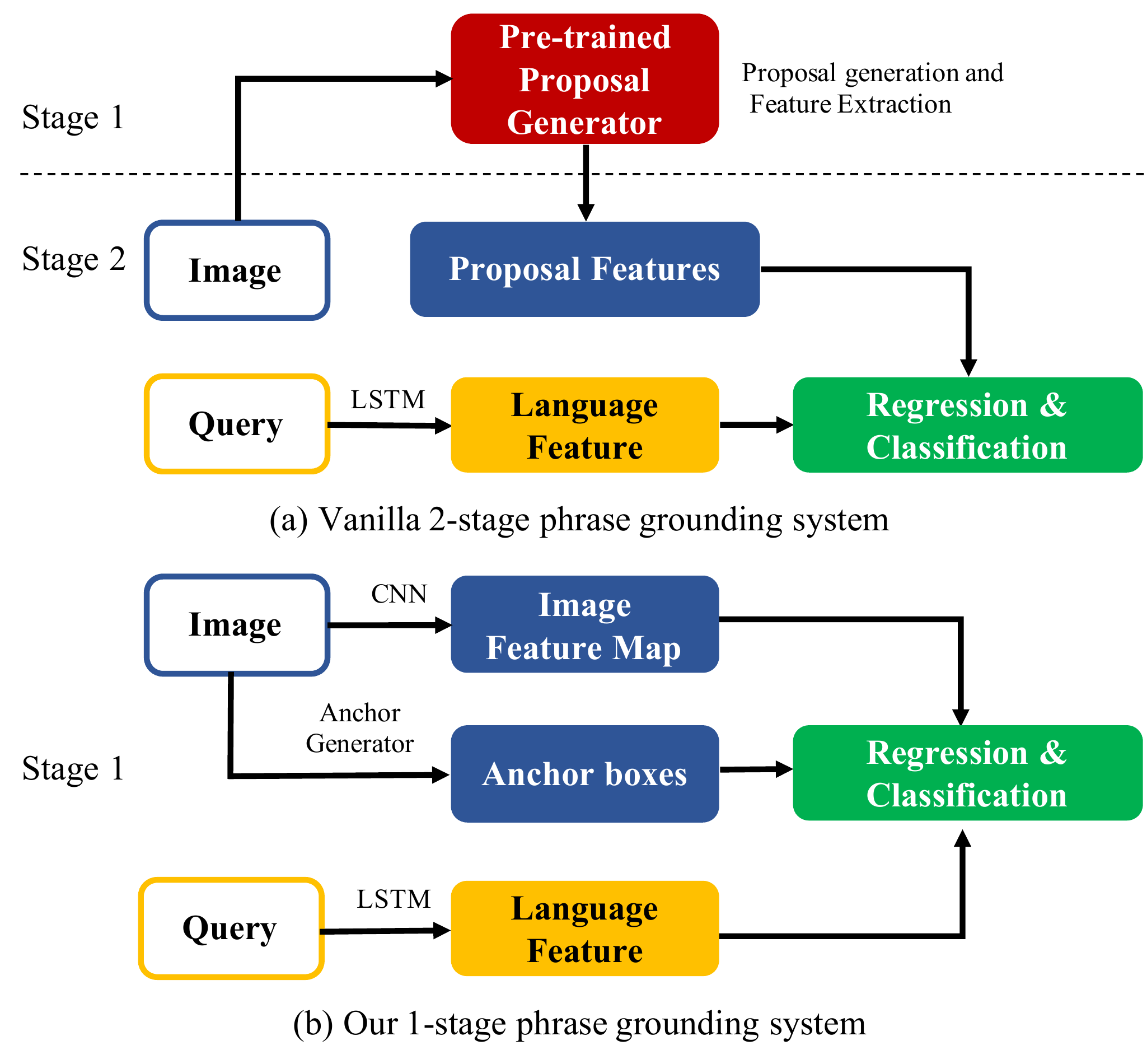}
    \caption{Previous phrase grounding systems (a) produce a small subset of proposals without considering the query restricting it to the entities of the detection network. Our system (b)  considers dense proposals, looks at the query to disambiguate and learns directly from grounding dataset}
    \label{fig:intro}
\end{figure}

\begin{figure*}[t]
    \centering
    \includegraphics[width=\linewidth]{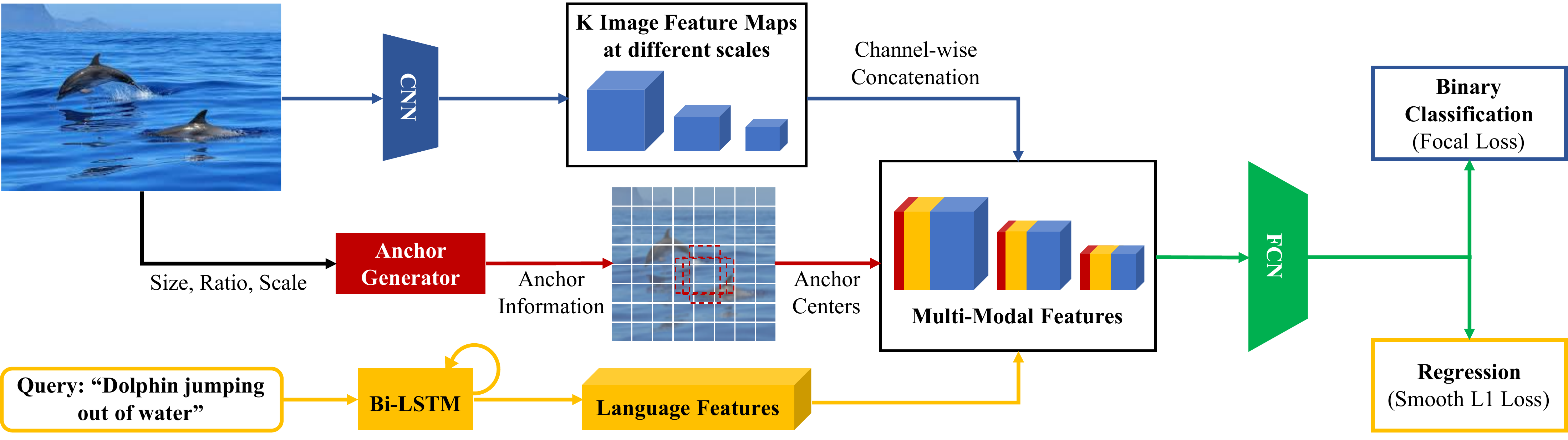}
    \caption{A schematic of the ZSGNet Architecture. Input to the system is an image-query pair. A deep network is used to produce $K$ image feature maps taken at different resolutions. The anchor generator uses the image size to produce anchors at different scales and resolution. We append the anchor centers at each cell of each feature map. The query phrase is encoded using a bidirectional LSTM (Bi-LSTM) and the language feature obtained is appended at every cell location of every feature map along the channel dimension. The resulting multi-modal feature maps are input to a Fully Convolution Network (FCN) block to output a prediction score and regression parameters which are trained using focal-loss ($L_{pred})$) and SmoothL1-loss($L_{reg})$) respectively.}
    \label{fig:arch}
\end{figure*}

We propose the following new formulation: Given an image $I$ with fixed candidate boxes (also called anchor boxes or dense proposals) $DP = [dp_1, \hdots dp_N]$  and a query phrase $P$, the task is to choose the best candidate box $dp_i$ and regress it to a tight bounding box $b_i$.
Since our candidate boxes depend only on the size of the image, we can use any image encoder to compute the visual features at run-time and remove the need for a pre-trained object detector as
illustrated in Fig \ref{fig:intro}.
This design is similar to single-shot architecture used in object detection \cite{liu2016ssd, lin2018focal}.

\textbf{Framework:} Our model consists of a language module to encode the query phrase, a visual module to obtain image feature maps followed by fully convolutional networks to output a $5d$ vector (for each candidate box) one for score and the rest for regressing to a tighter bounding box.
Fig \ref{fig:arch} provides an overview of our proposed architecture.

ZSGNet directly learns about the entities in a grounding dataset in an end-to-end fashion.
Moreover, since a query phrase refers to a particular object with possibly different attributes, the visual features are no longer invariant to intra-class changes.
This way, we address the limitations posed by previous systems.
Finally, owing to its single-stage approach, the network is computationally efficient.

 \textbf{Language module} consists of an embedding layer followed by a Bi-LSTM \cite{hochreiter1997long, schuster1997bidirectional} to encode the input query phrase.
 Given a query phrase $P = \{w_i\}_{i=1}^n$ we use GloVe vectors \cite{pennington2014glove} to encode each word in $P$ as word embedding vectors $\{\mathbf{w}_i\}_i^n\in\mathbb{R}^{d_q}$, where $d_q$ is the dimension of the embedding vector.
 These are fed into a Bi-LSTM \cite{hochreiter1997long,schuster1997bidirectional}.
 We use the normalized last hidden state vector $\{\hat{h}\}\in\mathbb{R}^{2d_l}$ of Bi-LSTM as the query feature, where $d_l$ is the dimension of the hidden layer of a single LSTM.

\textbf{Visual Module} consists of an image encoder to produce $K$ feature maps $\{v_i\}_{i=1}^K$ at different resolutions. We use ResNet-50 \cite{he2016deep} with FPN \cite{lin2017feature} as our default image encoder. We first normalize the visual feature maps along the channel dimension. Then we expand the language feature to the same dimensions of the visual feature maps and concatenate it along the channel dimension for each normalized visual feature map $\hat{v}_i$.
Finally, we append the normalized locations of the feature maps ($C_x, C_y = [c_x / W, c_y / H]$) along the channel dimension to aid in location based grounding (phrases which contain location information) and obtain the multi-modal feature maps $m_i$. At a  particular index of the $i^{th}$ feature map (indexed by $x, y$) we have
\begin{equation}
    m_i[x, y] = [\hat{v}_i[x, y] ; \hat{h} ; C_x ; C_y]
\end{equation}
where $;$ denotes the concatenation operation.

\textbf{Anchor Matching}
Following \cite{lin2018focal} we match 9 candidate boxes to every index of a feature map.
We use a fully convolutional network (FCN) to process the multi-modal features to output a 5 dimensional output (score and regression parameters) for each box.
In the anchor matching step, we use an $IoU$ threshold of $0.5$ (found ideal via experimentation).

\textbf{Loss Function}
For the binary classification into fore-ground and background we use the focal loss as described in \cite{lin2018focal}. For regressing to a tighter box we use the same encoding scheme as \cite{ren2015faster} with smooth-L1 loss.

Let $dp_j$ denote the $j^{th}$ anchor and $gt$ denote the ground truth bounding box. Let
\begin{align}
    g_{dp_j} &= \mathbb{1}_{IoU(dp_j, gt) \ge 0.5} \\
    G &= \{g_{dp_j} | g_{dp_j} = 1\}
\end{align}

Here $\mathbb{1}$ denotes the indicator random variable.
Thus, $g_{dp_j} = 1$ means the candidate box $dp_j$ matches with the ground-truth box and $G$ is the set of all such candidate boxes.
Now denoting focal loss \cite{lin2018focal} with default parameters ($\alpha = 0.25$,$\gamma=2$) by $L_F$ and the predicted score for the box $dp_j$ as $p_{dp_j}$ we get
\begin{equation}
    L_{pred} = \frac{1}{|G|} \sum_{j=1}^{|DP|} L_F(p_{dp_j}, g_{dp_j})
\end{equation}

Similarly, denoting SmoothL1-loss by $L_S$, the predicted regression parameters by $r_{dp_j}$  and the ground-truth regression parameters by $gt_{dp_j}$ we get
\begin{equation}
    L_{reg} = \frac{1}{|G|}\sum_{j = 1}^{|DP|} g_{dp_j} L_S(r_{dp_j}, gt_{dp_j})
\end{equation}
The final loss is calculated as $L = L_{pred} + \lambda L_{reg}$.
Here $\lambda$ is a hyper-parameter (we set $\lambda=1$).

\textbf{Training:} We match the candidate boxes (anchors) to each feature map generated by the feature encoder.
 We classify each candidate box as a foreground or a background using a prediction loss ($L_{pred}$) and regress it to get a tighter box ($L_{reg}$).
Foreground means that the candidate box shares $IoU \ge 0.5$ with the ground truth box.
For the regression loss we only consider the foreground candidate boxes.

\textbf{Testing} At test time, we choose the candidate box with the highest score and use its regression parameters to obtain the required bounding box.

\section{Experiments}
This section describes the dataset construction methods, followed by experiments and visualization.

\subsection{Dataset Construction}
\label{s:dc}
We sub-sample Flickr30k  \cite{plummer2015flickr30k} and Visual Genome \cite{krishna2017visual} to create datasets for the cases described in Section \ref{s:zsg_exp}.

\textbf{Flickr30k Entities} contains 5 sentences per image with every sentence containing 3.6 queries and has bounding box information of the referred object and its category (\eg ``people'', ``animal'').

\textbf{Visual Genome (VG)} has a scene graph for every image. The objects in the scene-graph are annotated with bounding boxes, region description and a synset (obtained from \cite{miller1995wordnet}).

We briefly describe the steps taken to create the ZSG datasets (more details can be found in the supplementary material). We follow the notation described in Table \ref{t:notation}.

\textbf{Case 0} is sampled from Flickr30k Entities\cite{plummer2015flickr30k}.
We need to ensure that $Q \notin W$.
For this, we first obtain lemmatized representation of the query words.
As the query phrases are  noun-phrases of a complete annotated sentence, the query word $Q$ referring to the noun  is almost always the last word of the query phrase $P$, we take it be so.
We do a 70:30 split of the extracted words to obtain ``included'' and ``excluded'' word lists respectively.
We then create a training set from the included list and  validation and tests from the excluded list, removing images that have overlap between train, validation or test lists.
We call the resulting split \textbf{Flickr-Split-0}.

\textbf{Case 1}
is also sampled from Flickr30k Entities \cite{plummer2015flickr30k} but this time we also the use predefined entity information.
We need the referred object $A$ to belong to a  category which is not in $C$.
Flickr30k has 7 common object categories  (\eg ``people'', ``animals'') and one category called ``other'' for objects which do not belong to the seven categories.
We take images with ``other'' objects and split them evenly to create validation and test sets,
The  remaining images comprise the train set; we remove any box annotations that belong to the ``other'' category to create \textbf{Flickr-Split-1}.

\textbf{Case 2 and Case 3} are sampled from Visual Genome \cite{krishna2017visual}. In addition to region phrases, visual genome also provides entity names mapped to synsets in wordnet \cite{miller1995wordnet}.
We count all the objects present in the dataset, choose $topI(=1000)$ objects and get their word embeddings, skipping without an embedding (we use GloVe \cite{pennington2014glove} trained on common crawl corpus).
We apply K-Means clustering ($K=20$) to cluster similar words.
We sort the words in each cluster $k$  by their frequency in the dataset and take the top half to be in the seen objects set ($S_k$) and the bottom half to be in the unseen objects set ($U_k$).
If an image contains an object $o_i$ such that $o_i \in U_k$ and another object $o_j \in S_k$ then it is an example of Case 3.
If no such $o_j$ exists then it is Case 2.
Finally, we take the union of images of the two cases to constitute the test set. We call the resulting splits \textbf{VG-Split-2} and \textbf{VG-Split-3}. This design ensures that for both \textbf{Cases 2 and 3} the referred object $A$ (in the test set) has a set of objects $B$ (in the training set) which are in the same semantic cluster.

We consider the remaining images to be candidates for the training set and include them if they contain at least one  object $o_i$ in S (where $S = \cup_k S_k$) and remove the annotations for any object $o_j \in U$ (where $U = \cup_k U_k$). This ensures that the training set contains objects in $S$ and does not contain any objects in $U$. However, such a training set turns out to be extremely imbalanced with respect to the clusters as clusters containing common entities such as ``person'' are much more prevalent than clusters containing ``cakes''.  We balance the training set following a simple threshold based sampling strategy (details in supplementary material) which results in most clusters (except $2$) to have similar number of query phrases. We follow the same strategy to create balanced test splits of \textbf{VG-Split-2} and \textbf{VG-Split-3}.

\textbf{Dataset Caveats:} (i) We note that polysemy is not taken care of \ie the same word can have different meanings.  (ii) Neither Visual Genome nor Flickr30k is a true referring expressions dataset \ie the query phrase may not always uniquely identify an object.


\subsection{Datasets Used}
\label{ss:dats}
\textbf{Flickr30k Entities} \cite{plummer2015flickr30k} contains 30k images each with 5 sentences and each sentences has multiple query phrases. We use the same splits used in \cite{chen2017query, hu2016natural, rohrbach2016grounding}.

\textbf{ReferIt(RefClef)} \cite{KazemzadehOrdonezMattenBergEMNLP14} is a subset of Imageclef \cite{escalante2010segmented} containing 20k images with 85k query phrases.
We use the same split as \cite{chen2017query, hu2016natural, rohrbach2016grounding}.

\textbf{Flickr-Split-0} We create an unseen split of Flickr30k based on the method outlined in Section~\ref{s:dc}. It contains 19K train images with 11K queries, 6K validation images with 9K queries and 6K test images with 9K queries.

\textbf{Flickr-Split-1} This split of Flickr30k has ``other'' category only in the validation and test images. It contains 19k training images with 87k query phrases and 6k images with 26k query phrases for validation and test each.

\textbf{VG-Split} We use a balanced training set (as described in Section \ref{s:dc}) containing 40K images and 264K query phrases. We use a subset (25\%) for validation. \textbf{VG-Split-2} contains 17K images and 23K query phrases in the unbalanced set, 10K images and 12K query phrases in the balanced set. \textbf{VG-Split-3} contains 41K images with 68K query phrases in the unbalanced set, 23K images and 25K query phrases in the balanced set.


\subsection{Experimental Setup}
\label{ss:exp_set}
\textbf{Evaluation Metric:}
We use the same metric as in \cite{chen2017query}. For each query phrase, we assume that there is only one ground truth bounding box. Given an image and a query phrase if the $IoU$ of our predicted bounding box and the ground truth box is more than $0.5$
we mark it as correct.
However, in the case of Visual Genome splits, the annotations are not precise so we use $0.3$ as the threshold.
The final accuracy is averaged over all image query phrase pairs.

\textbf{Baselines:}
To explicitly compare the performance of dense proposals, we create a new baseline \textbf{QRG} based on QRC \cite{chen2017query} which uses GloVe embeddings instead of embeddings learned from the data.
We benchmark it on Flickr30k to show there is no drop in performance compared to QRC.
We further use it as a strong baseline on the unseen splits.
In all cases, we use a fasterR-CNN \cite{ren2015faster} pretrained on Pascal-VOC \cite{everingham2010pascal} and fine-tune it on the target dataset.
For \textbf{Flickr-Split-0} we fine-tune on all the entities, for \textbf{Flickr-Split-1} we fine-tune on all entities except ``other''.
We use the top-100 box predictions provided by the fine-tuned network after applying non-maxima suppression to be consistent with implementation in \cite{chen2017query}.
For \textbf{VG-Split}, we train on all the seen-classes, \ie union of all seen objects in every cluster ($\cup_{k} S_k$). In this case, we don't use non-maxima suppression and instead consider all the 300 boxes provided by the fine-tuned region-proposal network.


\textbf{Implementation details:}
We train ZSGNet and baseline models till validation accuracy saturates and report our values on the test set. We  found Adam \cite{kingma2014adam} with learning rate $1e^{-4}$ for 20 epochs to be sufficient in most cases.
For ZSGNet, to generate image feature maps at different scales, we use two variations: (i) SSD \cite{liu2016ssd} with VGG network (ii) RetinaNet \cite{lin2018focal} with Resnet-50 \cite{he2016deep} network.
Note that these are not pretrained on any detection dataset.
Initially, we resize the image to $300 \times 300$ for faster training and later retrain with image sizes $600 \times 600$ which gives a consistent $2-3\%$ improvement.
We note that while image augmentations (like horizontal flipping) are crucial for object detectors it is harmful for grounding as the query phrases often have location information (like ``person standing on the left'', ``person to the right of the tree'').

\subsection{Results on Existing Grounding datasets}
\label{ss:res_exgrd}
\begin{table}[t]
\centering
\begin{tabular}{|c|c|c|c|}
\hline
Method & Net & Flickr30k & ReferIt \\ \hline
SCRC \cite{hu2016natural} & VGG & 27.8 & 17.9 \\ \hline
GroundeR (cls) \cite{rohrbach2016grounding} & VGG & 42.43 & 24.18 \\ \hline
GroundeR  (det) \cite{rohrbach2016grounding} & VGG & 48.38 & 28.5 \\ \hline
MCB (det) \cite{fukui2016multimodal} & VGG & 48.7 & 28.9 \\ \hline
Li (cls) \cite{li2017deep} & VGG & - & 40 \\ \hline
QRC* (det) \cite{chen2017query} & VGG & 60.21 & 44.1 \\ \hline
CITE* (cls) \cite{plummer2018conditional} & VGG & 61.89 & 34.13 \\ \hline
QRG* (det)  & VGG & 60.1 & - \\ \hline
\textbf{ZSGNet (cls)} & \textbf{VGG} & \textbf{60.12} & \textbf{53.31} \\ \hline
\textbf{ZSGNet (cls)} & \textbf{Res50} & \textbf{63.39} & \textbf{58.63} \\ \hline
\end{tabular}
\vspace{2mm}
\caption{Comparison of our model with other state of the art methods. We denote those networks which use classification weights from ImageNet \cite{russakovsky2015imagenet} using ``cls" and those networks which use detection weights from Pascal VOC \cite{everingham2010pascal} using ``det". The reported numbers are all $Accuracy@IoU=0.5$ or equivalently $Recall@1$. Models marked with ``*" fine-tune their detection network on the entities in the Flickr30k.}
\label{t1:sota}
\end{table}

\begin{table*}[t]
\centering
\begin{tabular}{|c|c|c|c|c|c|c|c|c|c|}
\hline
Method & Overall & people & clothing & bodyparts & animals & vehicles & instruments & scene & other \\ \hline
QRC - VGG(det) & 60.21 & 75.08  & 55.9     & 20.27     & 73.36   & 68.95    & 45.68       & 65.27 & 38.8  \\ \hline
CITE - VGG(det) & 61.89 &  \textbf{75.95}  & 58.50     & 30.78     & \textbf{77.03}   & \textbf{79.25}    & 48.15       & 58.78 & 43.24 \\ \hline
ZSGNet - VGG (cls) & 60.12  & 72.52	& 60.57	& 38.51	& 63.61 & 64.47	& 49.59 & 64.66	& 41.09 \\ \hline
ZSGNet - Res50 (cls) & \textbf{63.39}  & 73.87  & \textbf{66.18}    & \textbf{45.27}     & 73.79   & 71.38    & \textbf{58.54}  & \textbf{66.49} & \textbf{45.53} \\ \hline
\end{tabular}
\vspace{2mm}

\caption{Category-wise performance with the default split of Flickr30k Entities.}
\vspace{2mm}
\label{t1:cwf}
\end{table*}

\begin{table*}[t]
\centering
\begin{tabular}{|c|c|c|c|c|c|c|c|c|c|c|c|}
\hline
\multirow{2}{*}{Method} & \multirow{2}{*}{Net} &
\multirow{2}{*}{\begin{tabular}[c]{@{}c@{}}Flickr-\\ Split-0\end{tabular}} &
\multirow{2}{*}{\begin{tabular}[c]{@{}c@{}}Flickr-\\ Split-1\end{tabular}} & \multicolumn{2}{c|}{VG-2B} & \multicolumn{2}{c|}{VG-2UB} & \multicolumn{2}{c|}{VG-3B} & \multicolumn{2}{c|}{VG-3UB} \\ \cline{5-12}
 &  &  &  & 0.3 & 0.5 & 0.3 & 0.5 & 0.3 & 0.5 & 0.3 & 0.5 \\ \hline
QRG & VGG & 35.62 & 24.42 & 13.17 & 7.64 & 12.39 & 7.15 & 14.21 & 8.35 & 13.03 & 7.52 \\ \hline
\multirow{2}{*}{ZSGNet} & VGG & 39.32 & 29.35 & 17.09 & 11.02 & 16.48 & 10.55 & 17.63 & 11.42 & 17.35 & 10.97 \\ \cline{2-12}
 & Res50 & \textbf{43.02} & \textbf{31.23} & \textbf{19.95} & \textbf{12.90} & \textbf{19.12} & \textbf{12.37} & \textbf{20.77} & \textbf{13.77} & \textbf{19.72} & \textbf{12.82} \\ \hline
\end{tabular} \vspace{2mm}
\caption{Accuracy across various unseen splits. For Flickr-Split-0,1 we use Accuracy with $IoU$ threshold of $0.5$. Since Visual Genome annotations are noisy we additionally report Accuracy with $IoU$ threshold of $0.3$. The second row denotes the $IoU$ threshold at which the Accuracy is calculated. ``B'' and ``UB'' denote the balanced and unbalanced sets.}
\label{t:unseen_splits}
\end{table*}

Table \ref{t1:sota} compares ZSGNet with prior works on Flickr30k Entities \cite{plummer2015flickr30k} and ReferIt \cite{KazemzadehOrdonezMattenBergEMNLP14}.
We use ``det'' and ``cls'' to denote models using Pascal VOC \cite{everingham2010pascal} detection weights and ImageNet \cite{deng2009imagenet, russakovsky2015imagenet} classification weights.
Networks marked with ``*'' fine-tune their object detector pretrained on Pascal-VOC \cite{everingham2010pascal} on the fixed entities of Flickr30k \cite{plummer2015flickr30k}.

However, such information is not available in ReferIt dataset which explains $\sim9\%$ increase in performance of ZSGNet over other methods. This shows that our model learns about the entities directly from the grounding dataset.

For Flickr30k we also note entity-wise accuracy in Table \ref{t1:cwf} and compare against \cite{chen2017query, plummer2018conditional}. We don't compare to the full model in \cite{chen2017query} since it uses additional context queries from the sentence for disambiguation.
As these models use  object detectors  pretrained on Pascal-VOC \cite{everingham2010pascal}, they have somewhat higher performance on classes that are common to both  Flickr30k and Pascal-VOC (``animals'', ``people'' and ``vehicles'').
However, on the classes like ``clothing'' and ``bodyparts'' our model shows much better performance;  likely because both ``clothing'' and ``bodyparts'' are present along with ``people'' category and so the other methods choose the ``people'' category. Such biases are not exhibited in our results as  our model  is category agnostic.

\subsection{Results on ZSG Datasets}
\label{r:unseen}

Table \ref{t:unseen_splits} shows the performance of our ZSGNet model compared to QRG on the four unseen splits described in Section \ref{s:dc} and \ref{ss:dats}.
Across all splits, ZSGNet shows $4-8\%$ higher performance than QRG even though the latter has seen more data (the object detector is pretrained on Pascal VOC \cite{everingham2010pascal}).
Next, we observe that the accuracy obtained on Flickr-Split-0,1 are higher than the VG-split likely due to larger variation of the referred objects in Visual Genome.
Finally, the accuracy remains the same across the balanced and unbalanced sets indicating the model performs well across all clusters as our training set is balanced.

\begin{table}[t]
\centering
\resizebox{\linewidth}{!}{%
\begin{tabular}{|c|c|c|c|c|c|c|}
\hline
\multirow{2}{*}{} & \multirow{2}{*}{Method} & \multicolumn{5}{c|}{Semantic Distances} \\ \cline{3-7}
 &  & 3-4 & 4-5 & 5-6 & 6-7 & 7-8 \\ \hline \hline
\multirow{4}{*}{\begin{tabular}[c]{@{}c@{}}\\ \\ VG\\ 2B\end{tabular}} & \# I-P & 310 & 1050 & 3543 & 5321 & 1985 \\ \cline{2-7}
 & \begin{tabular}[c]{@{}c@{}}QRG\\ (Vgg)\end{tabular} & 25.16 & 16.67 & 15.16 & 10.96 & 12.54 \\ \cline{2-7}
 & \begin{tabular}[c]{@{}c@{}}ZSGNet \\ (Vgg)\end{tabular} & 28.71 & 21.52 & 19.02 & 15.37 & 14.76 \\ \cline{2-7}
 & \begin{tabular}[c]{@{}c@{}}\textbf{ZSGNet} \\ (\textbf{Res50})\end{tabular} & \textbf{31.94} & \textbf{25.14} & \textbf{21.99} & \textbf{17.89} & \textbf{17.98} \\ \hline \hline
\multirow{4}{*}{\begin{tabular}[c]{@{}c@{}}\\ \\ VG\\ 3B\end{tabular}} & \# I-P & 974 & 3199 & 7740 & 9873 & 3765 \\ \cline{2-7}
 & \begin{tabular}[c]{@{}c@{}}QRG\\ (Vgg)\end{tabular} & 23.1 & 20.13 & 14.73 & 12.19 & 11.24 \\ \cline{2-7}
 & \begin{tabular}[c]{@{}c@{}}ZSGNet\\ (Vgg)\end{tabular} & 23.82 & 25.73 & 17.16 & 16 & 14.56 \\ \cline{2-7}
 & \begin{tabular}[c]{@{}c@{}}\textbf{ZSGNet} \\ \textbf{(Res50)}\end{tabular} & \textbf{29.57} & \textbf{27.85} & \textbf{21.3} & \textbf{18.77} & \textbf{16.71} \\ \hline
\end{tabular}%
}\vspace{2mm}
\caption{Accuracy of various models on the balanced VG-Splits-2,3 w.r.t the semantic distance of the referred object ($A$) to the closest object seen at train time. VG-2B and VG-3B refer to the balanced test set for Case2, 3. \#I-P denotes the number of image-phrase pairs in the given semantic distance range.}
\label{t:sda}
\end{table}

We also study the relationship between accuracy and the distance of the referred object at test time ($A$) from the training set.
We re-use the clusters obtained while creating the \textbf{VG-Split} and consider the closest-seen object which lies in the same cluster as that of the referred object.
For every unseen object ($o_i$) in a particular cluster $k$ ($o_i \in U_k$) we find the closest seen object in the same cluster $o_j = \argmin_{o_j \in S_k} dist(o_i, o_j)$.
For $dist$ calculation, we use the GloVe embeddings \cite{pennington2014glove} corresponding to the objects and take the L2-norm of their difference.
We group the distances into 5 intervals of unit length and report the accuracy of the subset
where the distance of the referred object from the training set lies in that interval in Table \ref{t:sda}.

We note few examples of various intervals. We use the notation ($A$, $b$) \ie a tuple of the referred object and the closest object in the same cluster.
Examples for the interval 3-4 are \{(bouquet, flower), (cupcake, cake), (tablecloth, curtain)\} and for 7-8 are \{(printer, paper), (notebook, pen), (tattoo, poster)\}.
As expected, the accuracy declines with the semantic distance but smoothly \ie there is no sudden drop in performance.
\begin{figure*}[t]
\centering
\begin{tabular}{ccccc}
 \includegraphics[width=0.15\linewidth,height=0.15\linewidth]{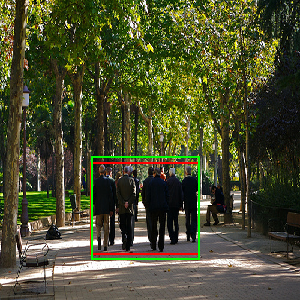} &
 \includegraphics[width=0.15\linewidth,height=0.15\linewidth]{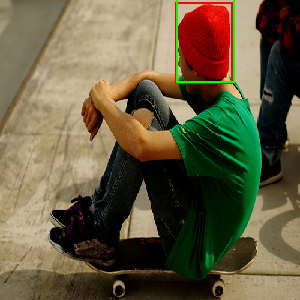} &
 \includegraphics[width=0.15\linewidth,height=0.15\linewidth]{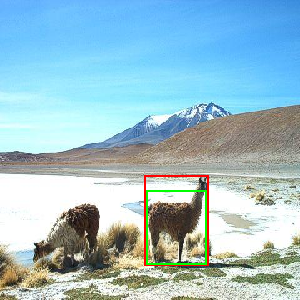} &
 \includegraphics[width=0.15\linewidth,height=0.15\linewidth]{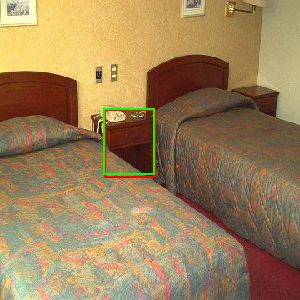} &
 \includegraphics[width=0.15\linewidth,height=0.15\linewidth]{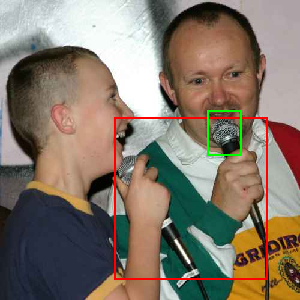} \\
a group of older men &  a red beanie cap & rightmost animal  & nightstand between beds & microphones \\

\includegraphics[width=0.15\linewidth,height=0.15\linewidth]{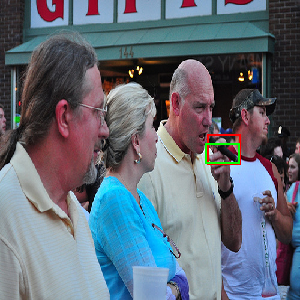} &
 \includegraphics[width=0.15\linewidth,height=0.15\linewidth]{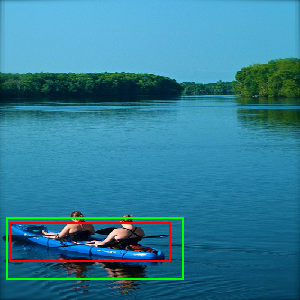} &
 \includegraphics[width=0.15\linewidth,height=0.15\linewidth]{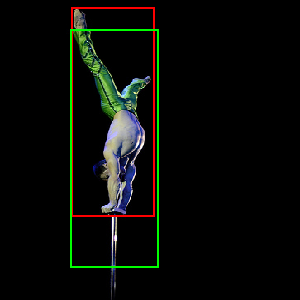} &
 \includegraphics[width=0.15\linewidth,height=0.15\linewidth]{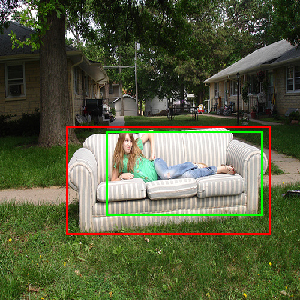} &
 \includegraphics[width=0.15\linewidth,height=0.15\linewidth]{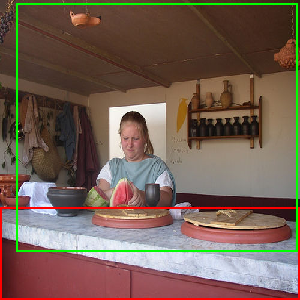}  \\
a cigar & a two-seat kayak &  a handstand  & couch & countertop \\

\includegraphics[width=0.15\linewidth]{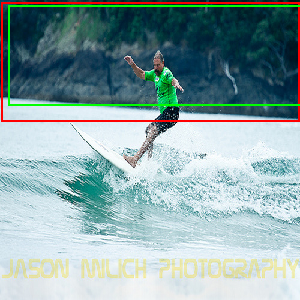}  &
\includegraphics[width=0.15\linewidth]{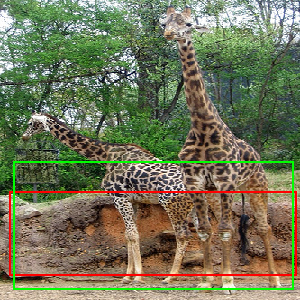} &
\includegraphics[width=0.15\linewidth]{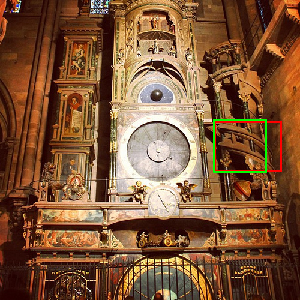} &
\includegraphics[width=0.15\linewidth]{images/fin_outs/VG_Case3_Test_5566.png} &
\includegraphics[width=0.15\linewidth]{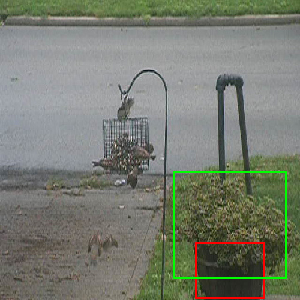} \\
a rocky \emph{cliff} (\emph{hill})
&
large  \emph{boulders} (\emph{rock})&
\emph{stairway} (\emph{wall})  &
\emph{shorts}  (\emph{person}) &
\emph{planter} (\emph{plant})
\\

\end{tabular}
\vspace{2mm}
\caption{Few grounding visualizations. In all cases, red denotes the ground truth box; green is the box predicted by ZSGNet. Row-1:Flickr30k, ReferIt; Row-2: Flickr-Split-0, 1; Row-3: VG-Split-2,3. In Row-3, the query word $Q$ is emphasised and the closest seen object is provided in parenthesis. The last column shows incorrect predictions.}

\label{f:flic}
\end{figure*}

\subsection{Ablation Study}
\label{ss:abl_study}
We show the performance of our model with different loss functions using the base model of ZSGNet on the validation set of ReferIt \cite{KazemzadehOrdonezMattenBergEMNLP14} in Table \ref{t:mabl}.
Note that using soft-max loss by itself places us higher than the previous methods.
Further using Binary Cross Entropy Loss and Focal loss \cite{lin2018focal} give a significant ($7\%$) performance boost which is expected in a single shot framework. Finally, image resizing
gives another $4\%$ increase.

\begin{table}[t]
\centering

\begin{tabular}{|c|c|}
\hline
Model             & Accuracy on RefClef \\ \hline
BM + Softmax      &   48.54          \\
BM + BCE      &     55.20            \\
BM  + FL          &    57.13           \\
BM + FL + Img-Resize & \textbf{61.75}\\ \hline
\end{tabular}
\vspace{2mm}
\caption{Ablation study: BM=Base Model, softmax means we classify only one candidate box as foreground, BCE = Binary Cross Entropy means we classify each candidate box as  the foreground or background, FL = Focal Loss, Img-Resize: use images of dimension $600 \times 600$}
\label{t:mabl}
\end{table}

\subsection{Visualization}
\label{ss:vis}
To qualitatively analyze our model we show a few visualizations in Fig \ref{f:flic}. The first row shows grounding results on \textbf{Flickr30k} (first, second column) and \textbf{ReferIt} (third, fourth column). Our model learns about the attribute(s) (``red''), location (``leftmost'') and entities (``cap'', ``nightstand'') and predicts very tight bounding box. In the last column our model incorrectly predicts only one ``microphone''.

The second row shows \textbf{Flickr-Split-0} (first, second column) and \textbf{Flickr-Split-1} (second, third column) predictions.
The query phrases ``cigar'', ``kayak'' are never encountered in the training set though close synonyms like ``cigarettes'', ``kayakers'' are \ie our model generalizes to unseen words even if they haven't been explicitly seen before.
This generalization can be attributed to pre-trained GloVe\cite{pennington2014glove} embedding.
On \textbf{Flickr-Split-1} the model predicts a good bounding box even though referred objects lies in a different category. However, when there are too many objects in the image the model gets confused (last column).

The third row shows predictions on \textbf{VG-Split-2} (first, second column) and \textbf{VG-Split-3} (third, fourth column). Additionally, we italicize the query word $Q$ which refers to the object $A$ in the image and mention the closest object encountered during training in parenthesis.
In \textbf{VG-Split-2} our model effectively utilizes word embedding knowledge and performs best when the closest object seen during training is visually similar to the referred object.
In \textbf{VG-Split-3} our model additionally needs to disambiguate between a seen object and the referred object and performs well when they are visually distinct like ``shorts'' and ``person''.
However, when the two are visually similar as in the case of ``planter'' and ``plants'' our model incorrectly grounds the seen object.

\section{Conclusion}
\label{s:conclusion}
In this work, we introduce the task of Zero-Shot grounding (ZSG) which aims at localizing novel objects from a query phrase. We outline four cases of zero-shot grounding to perform finer analysis.
We address the limitations posed by previous systems and propose a simple yet effective architecture ZSGNet.
Finally, we create new datasets by sub-sampling existing datasets to evaluate each of the four grounding cases. We verify that our proposed  model ZSGNet performs significantly better than existing baseline in the zero-shot setting.


\textbf{Acknowledgment:}
We thank Jiyang Gao for helpful discussions and the anonymous reviewers for helpful suggestions.
This paper is based
, in part,
on research sponsored by
the Air Force Research Laboratory and the Defense Advanced Research Projects Agency under agreement number FA8750-16-2-0204. The U.S. Government is authorized
to reproduce and distribute reprints for Governmental purposes notwithstanding any copyright notation thereon. The
views and conclusions contained herein are those of the authors and should not be interpreted as necessarily representing the official policies or endorsements, either expressed
or implied, of the Air Force Research Laboratory and the
Defense Advanced Research Projects Agency or the U.S.
Government.


{\small
\bibliographystyle{ieee_fullname}
\bibliography{egbib}
}

\clearpage
\appendix

\vskip .1in
\begin{center}
  {\Large \bf Appendix \par}
\end{center}
\renewcommand{\thesection}{\Alph{section}}

In this supplementary document, we present some of the details which couldn't be fit in the main paper. We provide details on how the datasets are sampled (Section \ref{s:supp-dc}) from Flickr30k Entities \cite{plummer2015flickr30k}, Visual Genome \cite{krishna2017visual} and their distributions (Section \ref{s:dd}).
We also provide (i) proposal recall of baseline method (Section \ref{s:pror}) (ii) image blind and language blind ablation of our model (Section \ref{s:blind_abl}).


\section{Dataset Construction}
\label{s:supp-dc}
We re-use the notation introduced in Table 1 of the main paper. We use Flickr30k for Case 0, 1 and Visual Genome for Case 2, 3 (reasons detailed in \ref{s:cd}).

\subsection{Case 0: $Q \notin W$}
This is sampled from \textbf{Flickr30k Entities} \cite{plummer2015flickr30k}.

In Flickr30k Entities each image has 5 associated sentences. The noun phrases (query-phrases) in each sentence are annotated with the bounding box information. Note that query-phrases in different sentences could refer to the same bounding box. Finally, each bounding box has an associated ``entity'' which we exploit in Case1.

For Case0, we consider the last word in the query phrase and use the lemmatized representation obtained from spacy \cite{spacy2}. This means that words like ``car'' and ``cars'' would be considered the same. However, this doesn't consider synonyms so ``automobile'' and ``cars'' are considered different.

We sort the lemmatized words in descending order of frequency and consider the $topI = 1000$ words to be always seen. This is reasonable for words like ``woman'', ``sky'' \etc.

Of the remaining words we do a 70:30 split and consider the first part to be in the include (seen) list ($S$) and the rest to be in the exclude (unseen) list ($U$). Note that even though $S, U$ are disjoint they could share few similar words.
The resulting include list contains $7k$ words and the exclude list contains $3k$ words.

For the test set we use only those images whose annotations have query word $Q \in U$. For the training set we consider the remaining image and remove annotations which have query word $Q \in U$. We also ensure that there is no overlap between the train and test images. The resulting split is called \textbf{Flickr-Split-0}.

The main motivation behind \textbf{Case0 Novel Words (NW)} is to see how well our model can perform without explicitly having seen the word during training.


\subsection{Case 1: $A \notin C$}
This is also sampled from \textbf{Flickr30k Entities} \cite{plummer2015flickr30k}. We use the entity information of each noun phrase (query-phrase) provided in the dataset. The entities provided are ``people'', ``clothing'', ``bodyparts'', ``animals'', ``vehicles'', ``instruments'', ``scene'' and ``other''. ``Other'' is used to denote those phrases which cannot be categorized into one of the remaining entities.

We extract out all the images with at least one phrase belonging to the ``other'' category. Of these, we randomly sample $50\%$ and use them for testing. Of the remaining images, we remove the annotations with the ``other'' category and use them for training.

The main motivation behind \textbf{Case1} is to see how well the model generalizes to novel object categories.

\subsection{Case 2, 3: $\exists B \text{ objects semantically close to } A$}
The two cases share the same training images but different test images. We sample the images and queries from the Visual Genome dataset \cite{krishna2017visual}. The dataset creation process has three major components: (i) cleaning the annotations to make them consistent (ii) clustering the objects and creating the train/test splits to satisfy the dataset properties of Case 2, 3 (iii) balancing the resulting splits.\\
\textbf{Cleaning Visual Genome Annotations:}
In visual genome each image has an average of 200 phrases. A phrase refers to a single object but may contain multiple objects in it.
Consider the phrase``man holding a pizza"; it is not directly specified if the referred object is a ``man" or a ``pizza" but there will be a bounding box in the image corresponding to the referred object, let us call it phrase BB; we need to infer the synset for this phrase  BB.  In addition, for each image, there are also annotated bounding boxes for each object type that appears in any of the phrases; in our example, there would be annotations for ``man", ``pizza" and other objects that may appear in other phrases.  To identify the synset for a phrase BB, we find the object bounding box that it has the maximum IoU with and use the object label associated with that bounding box.

Another difficulty is that if the same object instance is referred to in different phrases, it will have a different phrase BB associated with it. For consistency, we choose one and apply to all phrases. In implementation, we apply a non-maxima suppression algorithm; even though, there are no scores associated with the boxes, the algorithm selects on among highly overlapping alternatives. This step  provides us with a consistent set of annotations.

Even though the resulting annotations are consistent, the annotations are still spatially imprecise. Due to this reason, we recommend measuring detection accuracy with  with $IoU$ threshold of $0.3$ instead of the more common value of $0.5$.

\begin{figure*}[t]
\centering
\begin{tabular}{ccc}
 \includegraphics[width=0.3\linewidth]{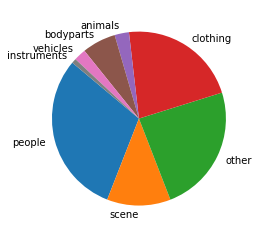} &
 \includegraphics[width=0.3\linewidth]{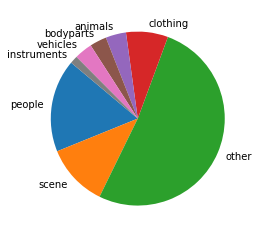} &
 \includegraphics[width=0.3\linewidth]{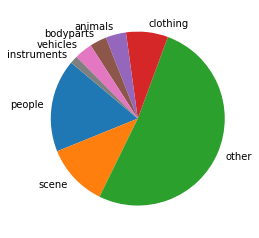} \\
 (a) Case0 Training Set &
 (b) Case0 Validation Set &
 (c) Case0 Test Set \\
  \includegraphics[width=0.3\linewidth]{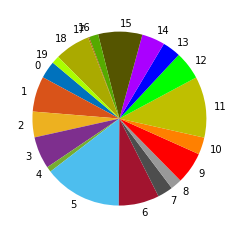} &
 \includegraphics[width=0.3\linewidth]{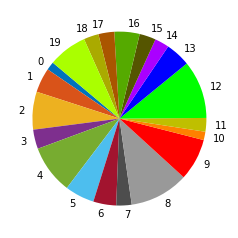} &
 \includegraphics[width=0.3\linewidth]{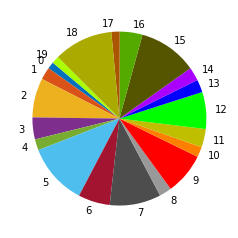} \\
 (d) Case2,3 Unbalanced Training Set &
 (e) Case2 Unbalanced Test Set &
 (f) Case3 Unbalanced Test Set \\
 \includegraphics[width=0.3\linewidth]{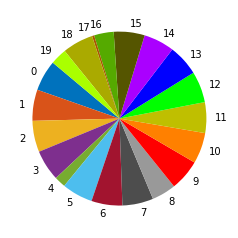} &
 \includegraphics[width=0.3\linewidth]{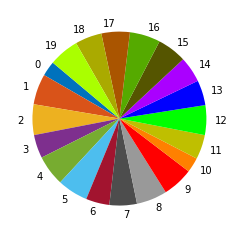} &
 \includegraphics[width=0.3\linewidth]{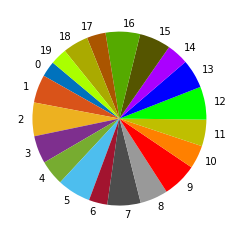} \\
 (g) Case2,3 Balanced Training Set &
 (h) Case2 Balanced Test Set &
 (i) Case3 Balanced Test Set \\
\end{tabular}
\vspace{2mm}
\caption{Category-wise distribution of various unseen splits. First row: training, validation and test set splits for Case 0; second row: unbalanced training and test sets for Case2 and Case 3; third row: balanced training and test sets for Case 2 and Case 3. In a row, the colors represent the same entities or the same clusters.}
\label{f1:splits}
\end{figure*}

\textbf{Clustering Objects: } Once we have a clean and consistent set of annotations, we sort all the objects (nearly $5k$ objects) by the number of appearances in the image. However, the objects at the tail end of the distribution are very infrequent so we consider only the top $1k$ objects.
Few of these don't have a corresponding word embedding (not available in spacy \cite{spacy2}) so we discard them.
This results in a total of $902$ objects.

Next, we cluster the GloVe \cite{pennington2014glove} word embeddings of the objects using K-Means clustering (with $K=20$).
We sort the objects in each cluster in descending order with respect to their frequency.
For a particular cluster $k$, we consider the first half to be ``seen'' ($S_k$) and the other half to be ``unseen'' ($U_k$). This gives us a total of $445$ seen objects and $457$ unseen objects. For a given cluster $k$ we consider all the images which have at least one object $o_i \in U_k$ to be test images. If there is another object in the same image $o_j$ such that $o_j \in S_k$, we put this image query pair into Case3 else into Case2.

For the remaining images, we remove annotations for any object $o_i \in \cup_{k}U_k$ and ensure there is at-least one object $o_i \in \cup_kS_k$ and use these to form the training set. However, by construction the training set turns out to be imbalanced with respect to clusters.

\textbf{Balancing the Dataset} To address the above issue we use the following balancing strategy:
  We use Zipf's law approximation that $freq \times rank \approx C$. That is as the rank of the cluster increases the number of annotations for that cluster decreases in a hyperbolic way. We use this to calculate an approximate mean of the clusters. Finally, we also consider $2 \times min\_cluster\_freq$ and take the max of the two.
  Thus, we have an approximate threshold at which we would like to sample. If for a particular cluster this threshold is more than the number of annotations in that cluster, we leave that cluster as it is, else we randomly sample $n = threshold$ annotations.
  Note that balancing is only done with respect to the clusters and not with respect to the object names.
Using this balancing strategy we get a balanced train set. We use $25\%$ of it for validation. For test sets we keep both balanced and unbalanced sets.

The main motivation for \textbf{Case2, 3} is to see how well the model generalizes to novel objects even if it depends on the semantic distance of the ``seen'' objects and if it can disambiguate the novel objects from the ``seen'' objects.

\begin{table}[t]
\centering
\resizebox{\linewidth}{!}{%
\begin{tabular}{|c|c|c|c|c|c|c|}
\hline
 & Flickr30k & ReferIt & \begin{tabular}[c]{@{}c@{}}Flickr\\ case0\end{tabular} & \begin{tabular}[c]{@{}c@{}}Flickr\\ case1\end{tabular} & \begin{tabular}[c]{@{}c@{}}VG\\ 2B\end{tabular} & \begin{tabular}[c]{@{}c@{}}VG\\ 3B\end{tabular} \\ \hline
FR (no f/t)& 73.4 & 25.4 & 64.95 & 62.9 & 15.87 & 13.92 \\ \hline
FR (f/t) & 90.85 & 58.35 & 85.18 & 74.85 & 26.17 & 25.07 \\ \hline
\end{tabular}%
}
\vspace{2mm}

\caption{Proposal Recall Rates using top-300 proposals at $IoU=0.5$ ($0.3$ for VG) calculated on test sets.
FR: FasterRCNN, no f/t: pretrained on pascal voc, f/t: fine-tuned on the target set. For referit we use f/t model on Flickr30k to be consistent with QRC.}
\label{tab:prop_recall}
\end{table}

\subsection{Choice of Datasets}
\label{s:cd}
We note that Flickr30k Entities doesn't provide synset information which is important to disambiguate synonym cases hence it cannot be used for Case2, 3.
Visual Genome doesn't contain wide categories like ``vehicles'' hence it cannot be used for Case 1.
For Case0, we could use Visual Genome as well, however, we choose Flickr30k Entities due to its precise bounding boxes.

\section{Dataset Distributions}
\label{s:dd}
We provide statistics for each dataset in Fig \ref{f1:splits}.
For Case0 we show the entity-wise distribution (a),(b),(c).
In particular we note that the ``other'' category occupies a larger set in the validation and test sets.
This is because the ``other'' category has a more diverse vocabulary and encompasses a larger part of the exclude vocabulary list.
For Case1, since it only has ``other'' category in its validation and test set, the entity-wise distributions are not meaningful and we don't include them here.

For Case2,3 we show the distributions with respect to the clusters formed via K-Means for both the unbalanced [(d),(e),(f)] and balanced cases [(g), (h), (i)]. We don't train on the unbalanced set but do test on the unbalanced set as well.
Note that the distribution across clusters in the balanced sets are uniform which means our balancing strategy was successful.

\section{Proposals from Pre-Trained Detector(s)}

\begin{table}[t]
\centering
\resizebox{\linewidth}{!}{%
\begin{tabular}{|c|c|c|c|c|c|c|}
\hline
Model
 & Flickr30k & ReferIt & \begin{tabular}[c]{@{}c@{}}Flickr\\ case0\end{tabular} & \begin{tabular}[c]{@{}c@{}}Flickr\\ case1\end{tabular} & \begin{tabular}[c]{@{}c@{}}VG\\ 2B\end{tabular} & \begin{tabular}[c]{@{}c@{}}VG\\ 3B\end{tabular} \\ \hline

LB & 0.008 & 0.0042 & 0.009 & 0.0024 & 0.0084 & 0.0093 \\ \hline
IB & 28.07 & 24.75 & 24.42 & 17.15 & 9.5 & 9.27 \\ \hline

\end{tabular}}
\vspace{2mm}

\caption{Ablation study: Language Blind (\textbf{LB}) and Image Blind (\textbf{IB}) setting using Images of Resolution $300 \times 300$. Metric reported is Accuracy@IoU=0.5 (0.3 for VG)}
\label{t:limgabl}
\end{table}

\label{s:pror}

A crucial difference between \textbf{ZSGNet} and prior work is the removal of proposals obtained from a pre-trained network.
To explicitly analyze the the errors caused due to missing proposals we calculate the proposal recall.

\textbf{Proposal Recall:} We measure the recall rates (@300) of the region proposal network (RPN) from FasterRCNN \cite{ren2015faster} pretrained on Pascal VOC \cite{everingham2010pascal} and fine-tuned on the target dataset in Table \ref{tab:prop_recall}. For ReferIt \cite{KazemzadehOrdonezMattenBergEMNLP14} we use the fine-tuned model on Flickr30k Entities
\cite{flickr30k} to be consistent with QRC \cite{chen2017query}. We note that (i) proposal recall significantly improves when we fine-tune on the target dataset (ii) performance of QRG on Flickr30k, case0, case1 follows the same trend as the proposal recall (iii) proposal recall is significantly smaller on Visual Genome \cite{krishna2017visual} due to (a) a large number of classes in visual genome (b) considering the ``unseen'' classes during training as negatives. These recall scores motivate the use of dense proposals for zero-shot grounding.

\section{Image Blind and Language Blind Ablations}
\label{s:blind_abl}

 \textbf{Model Ablations:} We ablate our model in two settings: language blind (\textbf{LB}) (the model sees only the image and not the query) and image blind (\textbf{IB}) (the model considers the query but not the image).
We provide the results obtained after retraining the model in Table \ref{t:limgabl}.
In the \textbf{LB} case, our model sees multiple correct solutions for the same image and therefore gives a random box output leading to a very low accuracy across all datasets. In the  \textbf{IB} case, our model learns to always predict a box in the center.
We note that the referred object lies in the center of the image for Flickr30k and ReferIt. This is because Flickr30k Entities contains queries derived from captions which refer to the central part of the image and ReferIt is a two player game with a high chance of referring to the central object, leading to relatively high accuracy $25-30\%$.

However, this is substantially lower for Visual Genome ($9-10\%$) which has denser object annotations.

\end{document}